\documentclass[utf8x]{article-hermes_frenchb}
\usepackage[labelfont=bf,textfont=it,labelsep=period,justification=raggedright,singlelinecheck=false]{caption}
\usepackage{stmaryrd}
\usepackage{braket, xcolor}

\journal{TAL. Volume 64 -- n°1/2023}{11}{35}

\title[Traitement quantique des langues]{Traitement quantique des langues : état de l'art}

\author{Sabrina Campano \andauthor Tahar Nabil \andauthor Meryl Bothua} 

\address{EDF Lab Paris-Saclay, boulevard Gaspard Monge, 91120 Palaiseau}

\resume{Cet article propose un état de l'art des travaux d'informatique quantique pour le traitement automatique des langues (TAL). 
Leur objectif est d'améliorer les performances des modèles actuels, et de mieux représenter des phénomènes linguistiques comme l'ambiguïté ou les dépendances à longue distance. Plusieurs familles d'approches sont présentées, dont les modèles symboliques diagrammatiques, et les réseaux de neurones hybrides. Ces travaux montrent que des expérimentations sont déjà possibles, et ouvrent des perspectives de recherche concernant le développement de nouveaux modèles et leur évaluation.
}

\abstract{Quantum Natural Language Processing : a review}{This article presents a review of quantum computing research works for Natural Language Processing (NLP). Their goal is to improve the performance of current models, and to provide a better representation of several linguistic phenomena, such as ambiguity and long range dependencies. Several families of approaches are presented, including symbolic diagrammatic approaches, and hybrid neural networks. These works show that experimental studies are already feasible, and open research perspectives on the conception of new models and their evaluation.}

\motscles{
informatique quantique,
traitement quantique des langues.
}

\keywords{
quantum computing,
quantum natural language processing.}

\begin{document}

\maketitlepage

\newcommand{\fakesentence}{Attention à ce que les figures et les tableaux ne débordent pas dans les marges. }
\newcommand{\fakeparagraph}{
\fakesentence
\fakesentence
\fakesentence
\fakesentence
\fakesentence
\fakesentence
}

\newcommand{\TAL}{traitement automatique des langues}

\newcommand{\CAD}{c'est-à-dire}
\newcommand{\COLL}{et collègues}
\newcommand{\PEX}{par exemple}
\newcommand{\POPP}{par opposition à}

\newcommand{\cad}{c.-à-d.}
\newcommand{\coll}{et~coll.}
\newcommand{\pex}{p.~ex.}
\newcommand{\popp}{p.~opp.}


\section{Introduction}

Depuis que l'idée a été émise notamment par le physicien Richard Feynman en 1981, l'informatique quantique intrigue, générant de nombreux efforts de recherche pour comprendre et réaliser son potentiel \cite{feynman1982simulating}.
En tirant profit de la superposition d'états et de l'intrication, deux propriétés fondamentales des systèmes quantiques, la promesse des ordinateurs quantiques est d'obtenir des avantages calculatoires par rapport aux machines dites \frquote{classiques}, en réduisant la complexité algorithmique de résolution de certains problèmes.
Par exemple, les algorithmes de \citeasnoun{shor1994algorithms} pour la factorisation des nombres entiers et de Harrow-Hassidim-Lloyd (HHL) pour la résolution d'un système linéaire \cite{harrow2009quantum} garantissent respectivement, par la preuve théorique, des accélérations exponentielles par rapport à toute contrepartie classique.
L'algorithme de recherche d'information de Grover apporte quant à lui un gain quadratique \cite{grover1996fast}.
Le développement de méthodes algorithmiques précède ainsi l'avènement de l'ordinateur quantique, la première réalisation expérimentale d'un algorithme quantique -- celui de Grover -- datant de 1998 \cite{chuang1998experimental}.
Si la notion d'avantage quantique est multiforme et reste difficile à mesurer \cite{ronnow2014defining}, les progrès significatifs récents des constructeurs d'infrastructures matérielles permettent dorénavant d'élargir l'informatique quantique expérimentale à plusieurs disciplines.
{\`A} titre illustratif, \citeasnoun{madsen2022quantum} ont ainsi mené des essais démontrant un avantage sur une tâche d'échantillonnage de bosons.

Dans le domaine du traitement automatique des langues (TAL), 
l'informatique quantique a fait émerger un nouveau champ disciplinaire, le traitement quantique des langues (TQL, ou \emph{ QNLP, Quantum Natural Language Processing} en anglais).
Les origines du TQL remontent aux travaux de \citeasnoun{coecke2010mathematical}, où le sens d'une phrase est calculé en utilisant la composition de produits tensoriels, et la première conférence dédiée a eu lieu en 2019, accueillant à la fois des contributions théoriques et expérimentales.
Si les méthodes classiques de TAL connaissent des succès spectaculaires sur de nombreuses tâches, \pex{} \emph{via} les modèles d'apprentissage profond, la perspective du TQL tient avant tout aux fondements théoriques de la discipline.

{\`A} terme, il s'agit en effet de trouver une façon de représenter et de traiter la langue pouvant dépasser les limites actuelles des approches classiques. Au-delà de l'accélération des calculs, certains travaux en TQL sont motivés par des approches symboliques, qui pourraient permettre de mieux prendre en compte des aspects comme l'ambiguïté, et la façon dont le sens d'une phrase est composé à partir de ses constituants. \citeasnoun{meichanetzidis2023grammar} voient ces approches symboliques plus transparentes que celles fondées sur les réseaux de neurones, qualifiés de boîtes noires. Selon \citeasnoun{correia2022quantum}, les approches à base de règles en TAL pourraient être réinstituées, car les calculs requis deviendraient plus efficaces avec des algorithmes quantiques.


Dans ce contexte, l'objectif de cette revue de l'état de l'art en TQL est de familiariser le lecteur de la communauté TAL avec cette discipline naissante.
Nous définissons ainsi les concepts clés de l'informatique quantique en section \ref{sec:concepts}.
Puis la section \ref{sec:matricesdensite} présente de nouveaux modèles classiques de langue s'appuyant sur ces notions.
Les modèles quantiques sont décrits en section \ref{sec:modeles}, avec d'une part les approches diagrammatiques et symboliques, et d'autre part les hybridations quantiques des réseaux de neurones classiques.
L'inclusion de ces derniers modèles, la structuration fondée sur les approches plutôt que sur les cas d'application, et les explications détaillées fournies dans ce document complètent ainsi la première revue de l’état de l’art sur le sujet proposée par \citeasnoun{wu2021natural}.
Nous détaillons également quelques expérimentations, rendues possibles par les progrès technologiques.
La comparaison rigoureuse de ces méthodes entre elles ou bien à des algorithmes classiques n'entre pas toutefois dans le champ de cette revue, le domaine étant encore jeune.
Cela nous amène à discuter sur les limites actuelles et sur les développements futurs du TQL en section \ref{sec:discussion}.
Enfin, nous concluons en section \ref{sec:conclusion}.

\section{Concepts} \label{sec:concepts}

Introduisons tout d'abord les concepts clés nécessaires à la compréhension des travaux de recherche en TQL.
D'après le premier postulat de l'informatique quantique, un système physique isolé (fermé) est décrit par un vecteur \emph{ket} $\ket{\psi}$ appartenant à un espace de Hilbert\footnote{Espace vectoriel muni d'un produit scalaire.} à valeurs complexes, appelé espace d'état \cite{nielsen2010quantum}.
Le transconjugué complexe $\ket{\psi}^{\dagger}$ de $\ket{\psi}$ est noté \emph{bra} $\bra{\psi}$, aussi appelé \textit{effet}, ainsi la quantité $\braket{\phi|\psi}$ est un produit scalaire tandis que l'opérateur $\ket{\psi}\bra{\psi}$ est une matrice de projection sur la direction $\ket{\psi}$.
Un qubit $\ket{\psi}$, ou \frquote{bit quantique}, est un vecteur dans un espace d'état de dimension 2, il s'écrit comme une combinaison linéaire $\ket{\psi}=\alpha\ket{0}+\beta\ket{1}$, où $\alpha,\beta\in\mathbb{C}$ et $(\ket{0},\ket{1})$ est une base de l'espace, $\ket{0}=(1,\,0)^{\intercal}$ et $\ket{1}=(0,\,1)^{\intercal}$.
Il s'agit de la plus petite quantité d'information d'un système quantique.
Contrairement à son équivalent classique qui ne prend que deux valeurs, 0 ou 1, le qubit est dans une \emph{superposition} de valeurs entre les états $\ket{0}$ et $\ket{1}$.
Par suite, l'espace d'état d'un système à $n$ qubits a pour dimension $N=2^n$ et est le produit tensoriel des $n$ espaces à un qubit, avec pour base canonique l'ensemble des $\{\ket{j_1}\otimes\dots\otimes\ket{j_n}|\,j_k\in\{0,1\},k=1,\dots,n\}$; le \textit{i}-ème vecteur de la base est $\ket{e_{i}}=(\delta_{1i},\dots,\delta_{Ni})^{\intercal}$ où $\delta_{ij}=1$ si et seulement si $i=j$, 0 sinon.
Toutefois, bien qu'un $n-$qubit encode $2^n$ coefficients, seuls $n$ bits d'information classique peuvent en être extraits.
En effet, tout qubit décomposé dans la base des $\ket{e_i}$ s'écrit $\ket{\psi}=\sum_{i=1}^{N}\alpha_i\ket{e_i}$, $\alpha_i\in\mathbb{C}$, et le postulat de la mesure indique qu'une observation du système aura pour résultat de le figer dans l'état $\ket{e_i}$ avec la probabilité $|\braket{e_i|\psi}|^2=|\alpha_i|^2$.
Cela implique en particulier la condition $\sum_i|\alpha_i|^2=1$: les qubits sont des vecteurs unitaires.

Outre la superposition, une seconde propriété des systèmes quantiques donne l'intuition qu'ils pourraient \frquote{mieux} traiter l'information que les systèmes classiques: c'est l'\emph{intrication}.
Un $n$-qubit est intriqué s'il n'est pas séparable, \cad{} s'il ne peut pas s'écrire comme produit tensoriel de qubits de plus petite dimension\footnote{C'est le cas fameux de l'état de Bell $\frac{1}{\sqrt{2}}(\ket{00}+\ket{11})$, que l'on ne peut écrire comme produit de deux qubits -- où $\ket{00}=\ket{0}\otimes\ket{0}$ et similairement pour $\ket{11}$.}.
Dans un système intriqué, les propriétés des qubits sont fortement corrélées, plus que ne peuvent l'être des particules classiques, ce qui ouvre \emph{a priori} des possibilités algorithmiques inacessibles sur ordinateur classique \cite{nielsen2010quantum}.

Enfin, l'évolution d'un système de $n$-qubits entre deux mesures est décrite par un opérateur linéaire $U$, de taille $2^n\times2^n$, tel que $\ket{\psi(t_0)}\rightarrow\ket{\psi(t)}=U(t,t_0)\ket{\psi(t_0)}$.
$U$ conserve les angles et les normes (opérateur unitaire) : $U^{\dagger}U=UU^{\dagger}=I$, toute matrice unitaire de taille $2\times2$ applique donc une transformation par rotation d'un qubit -- voir \citeasnoun{nielsen2010quantum} pour une liste des opérateurs les plus fréquents (portes NON ou contrôlée NON, matrices de Pauli, porte d'échange, etc.).

Plus généralement, un système physique peut être soit dans l'état $\ket{\psi}$ (état pur, connaissance certaine du système), soit dans un mélange statistique classique d'états quantiques, \cad{} dans l'état $\ket{\psi_i}$ avec la probabilité $\theta_i\in[0,1]$, $\sum_i\theta_i=1$ (connaissance incomplète du système) -- chaque $\ket{\psi_i}$ pouvant lui-même être dans une superposition d'états quantiques.
Ces systèmes sont décrits par le formalisme des \emph{matrices de densité}, plutôt que par des vecteurs d'état:
pour un état pur, il s'agit de la projection $\rho=\ket{\psi}\bra{\psi}$, tandis qu'elle s'écrit $\rho=\sum_i\theta_i\ket{\psi_i}\bra{\psi_i}$ pour un mélange.
On montre que tout opérateur $\rho$ hermitien, positif\footnote{Opérateur hermitien: $\rho^{\dagger}=\rho$, positivité de $\rho$: pour tout $\ket{\psi}$, $\bra{\psi}\rho\ket{\psi}\geq0$.}, de trace $\text{tr}(\rho)=1$ est un opérateur de densité d'un système $\{\theta_i,\ket{\psi_i}\}$, et réciproquement -- cette équivalence définit l'opérateur de densité et permet de reformuler les postulats quantiques précédemment décrits \cite{nielsen2010quantum}.
L'entropie de Von Neumann $S(\rho):=-\text{tr}(\rho\log\rho)$ mesure l'incertitude associée au mélange, elle est nulle pour un état pur.
Enfin, $\rho$ attribue à tout événement $\ket{v}$ la probabilité $\text{tr}(\rho\ket{v}\bra{v})$, et généralise les distributions de probabilité classiques, qui correspondent elles à $\rho$ diagonale:
pour une loi classique de probabilité $\theta_i$ associée à l'événement $i$, il suffit d'associer le vecteur $\ket{e_i}$ de la base canonique à $i$ et de considérer $\rho_{\theta}=\sum_i\theta_i\ket{e_i}\bra{e_i}$.
$\rho_{\theta}$ est alors diagonale, et on vérifie que $\text{tr}(\rho_{\theta}\ket{e_i}\bra{e_i})=\theta_i$.
Nous verrons en section \ref{sec:matricesdensite} que de nombreux travaux de TQL s'appuient sur le formalisme général des matrices de densité pour encoder des relations de dépendance entre mots et pour élaborer de nouveaux modèles.

\begin{figure}
\centering
\includegraphics[width=.875\textwidth]{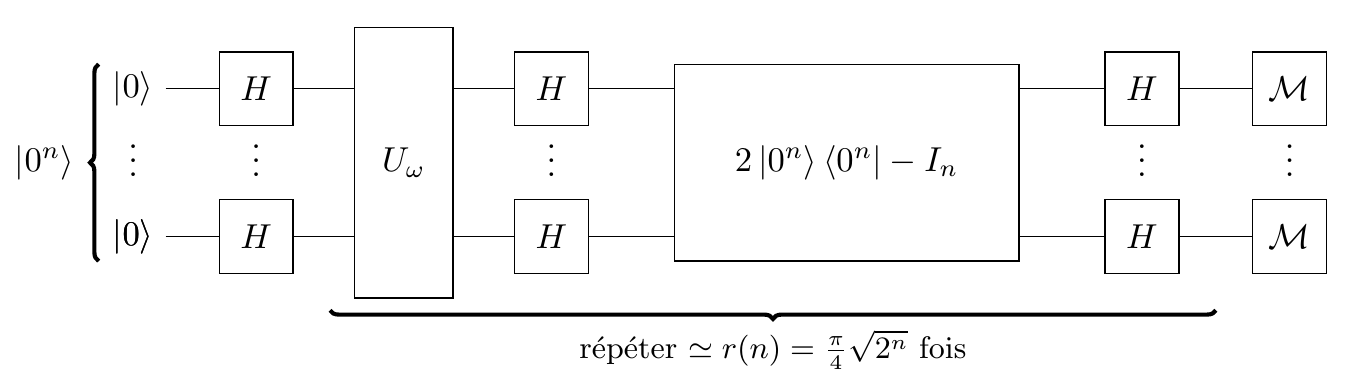}
\caption{Circuit quantique réalisant l'algorithme de \protect\citeasnoun{grover1996fast} pour la recherche non structurée d'information: trouver l'unique $\omega\in\{0,1\}^n$ tel que $f_{\omega}(x)=\delta_{x\omega}$ est non nulle, $x\in\{0,1\}^n$.
Un registre de $n$ qubits dans l'état initial $\ket{0^n}=\ket{0}\otimes\dots\otimes\ket{0}$ est mis dans la superposition de tous les qubits de la base canonique par la porte $H$ de Hadamard.
Une série d'opérateurs est répétée $r(n)$ fois, et la mesure finale donne l'état $\ket{\omega}$ avec probabilité au moins $1-4/2^n$.
}
\label{fig:grover}
\end{figure}

Les notions élémentaires étant établies, décrivons finalement le \emph{circuit quantique}, composant au c\oe ur des ordinateurs quantiques.
Un circuit quantique est un système fermé composé (i) d'un registre de $n$ qubits, typiquement préparés dans l'état $\ket{0}\otimes\dots\otimes\ket{0}$, (ii) d'une séquence d'opérateurs unitaires, les portes quantiques, modifiant l'état du système (iii) et enfin d'une action de mesure de l'état final du système \cite{divincenzo2000physical}.
Pour illustration, le circuit réalisant l'algorithme de Grover est représenté en figure \ref{fig:grover}.
La mesure de l'état final étant par nature aléatoire, l'exécution du circuit est répétée plusieurs fois -- on parle de \frquote{tirs} -- pour obtenir la probabilité du résultat.
Parmi les circuits, un \emph{circuit variationnel} (VQC, \emph{Variational Quantum Circuit}) est un circuit quantique paramétrique, \cad{} contenant des portes paramétrisées $U(\theta)$.
Les paramètres libres $\theta$ des portes du VQC peuvent être optimisés pour minimiser une fonction objectif: 
typiquement, un algorithme hybride optimisera de façon classique les paramètres du VQC, \pex{} par descente de gradient, avec une fonction coût fournie par la mesure du circuit.
Enfin, un \emph{ansatz} désigne une sous-séquence d'un VQC correspondant à une certaine architecture et remplissant une certaine fonctionnalité :
préparation de l'état initial, combinaisons spécifiques de portes paramétriques, etc.
Il s'agit de blocs élémentaires paramétriques à utiliser pour configurer le circuit, à la manière des blocs constitutifs des réseaux neuronaux classiques (type de couche neuronale, profondeur, etc.).
La configuration de l'\textit{ansatz} est un hyperparamètre du VQC et reste donc fixe lors de l'optimisation.

Notons cependant que tout circuit n'est pas implémentable en pratique.
En effet, le développement technologique actuel des ordinateurs quantiques reste contraint par le bruit quantique; c'est l'ère du NISQ, \emph{Noisy Intermediate Scale Quantum}, qui ne garantit pas que le système reste fermé à l'environnement extérieur.
Le NISQ induit une double limitation des circuits quantiques, la première en espace, en restreignant le nombre de qubits simultanément contrôlables (de l'ordre de 100 sur les machines actuelles), la seconde en temps, en réduisant la longueur des circuits pour maintenir la cohérence des états.
En outre, les machines quantiques actuelles ne sont pas capables d'implémenter toutes les portes, bien que des résultats d'universalité exacte ou approchée permettent de décomposer toute opération unitaire à $n$ qubits en un circuit composé d'un nombre fini de portes universelles \cite{nielsen2010quantum}.
Cela ajoute ainsi une étape de compilation des circuits afin de les convertir en portes universelles admises par la machine, ce qui rallonge leur longueur effective.

\section{Modèles de langue d'inspiration quantique} \label{sec:matricesdensite}

La théorie quantique a d'abord été employée pour produire de meilleurs modèles classiques de résolution de tâches fondamentales de TAL, sans recourir à du matériel quantique:
l'objectif est soit de mieux représenter les dépendances entre mots, soit de mieux représenter certaines relations lexicales comme l'homonymie ou l'hyponymie.

\subsection{Modéliser les dépendances entre termes: le modèle de langue quantique} \label{sssec:qlm}

Le modèle de langue quantique (QLM, \emph{Quantum Language Model}) de \citeasnoun{sordoni2013modeling} répond au premier objectif.
Soit un vocabulaire $\mathcal{V}$ de taille $n$,
le QLM associe un vecteur \emph{one-hot} $\ket{e_w}$ à tout mot $w\in\mathcal{V}$, par la projection $\Pi_{w}=\ket{e_w}\bra{e_w}$ sur le vecteur $\ket{e_w}$ de la base canonique de l'espace d'état de dimension $n$.
Les relations de cooccurrence entre $K$ termes $\{w_k\}_{k=1}^K$, ici, toute apparition des $\{w_k\}$ dans un ordre quelconque dans une fenêtre de taille fixe,
sont modélisées par la superposition \\ $\Pi_{\kappa}=\ket{\kappa}\bra{\kappa}$ (état pur), $\ket{\kappa}=\sum_{k=1}^K \lambda_k\ket{e_{w_{k}}}$, $\lambda_k\in\mathbb{R},\,\sum_k\lambda_k^2=1$.
Les $\lambda_k$ sont fixés, de façon uniforme ou d'après la fréquence inverse de document.
Prenons par exemple $\mathcal{V}=$ \emph{\{traitement, automatique, langues\}}, avec une cooccurrence des termes \emph{traitement} et \emph{langues} modélisée par la superposition $\ket{\kappa_{tl}}=\frac{1}{\sqrt{5}}\ket{e_{traitement}}$ $+\frac{2}{\sqrt{5}}\ket{e_{langues}}$.
Les projections $\{\Pi_i\}$ individuelles et de cooccurrence sont alors respectivement:
$$
\Pi_{t}=\begin{bmatrix}1 & 0 & 0 \\ 0&0&0\\0&0&0\end{bmatrix},\,\Pi_{a}=\begin{bmatrix}0 & 0 & 0 \\ 0&1&0\\0&0&0\end{bmatrix},\,\Pi_{l}=\begin{bmatrix}0 & 0 & 0 \\ 0&0&0\\0&0&1\end{bmatrix},\,\Pi_{tl}=\begin{bmatrix}\frac{1}{5} & 0 & \frac{2}{5}\\ 0 & 0 & 0\\ \frac{2}{5}& 0 & \frac{4}{5}\end{bmatrix}.$$
Un document $d$ est ainsi décrit par l'ensemble des projections $\{\Pi_i\}$ associées à ses termes individuels et aux sous-ensembles de mots cooccurrents --
sans ces dépendances, le QLM reviendrait à une loi classique sur les termes individuels (densité diagonale, section \ref{sec:concepts}).
L'ajout des événements de superposition permet de contrôler à quel point l'observation de la cooccurrence informe sur les constituants individuels, \textit{via} $\lambda_k$.
Enfin, la matrice de densité $\rho_d$ de taille $n\times n$ représentant $d$ est apprise à partir des $\Pi_i$ par maximum de vraisemblance: $\rho_d=\arg\max_\rho\sum_i\log\text{tr}(\rho\Pi_i)$, telle que $\rho$ est positive et de trace unitaire.
Ce problème difficile d'optimisation est résolu par l'algorithme itératif EM (\textit{expectation-maximization}), sans garantir la convergence rapide vers un optimum global. 

\citeasnoun{sordoni2013modeling} appliquent le QLM à des tâches de recherche d'information, consistant à trouver la réponse qui correspond le mieux à une requête utilisateur.
Après modélisation des documents du corpus, le modèle $\rho_q$ d'une requête est comparé au modèle $\rho_d$ de chaque document d'après l'entropie relative quantique $-\text{~tr}(\rho_q(\log\rho_q-\log\rho_d))$, qui généralise la divergence de Kullback-Leibler et classe donc les documents par degré de proximité avec la requête.
Pour conserver un temps d'exécution raisonnable\footnote{Il y a $2^{|Q|}$ sous-ensembles de mots possibles pour une requête de taille $|Q|$.}, le nombre de mots contenus dans la requête est limité à 3.
Les expérimentations sont effectuées avec le moteur de recherche \textit{open source} Indri sur 4 jeux de données de la collection TREC, contenant de 90 257 à 50 220 423 documents.
Le modèle QLM atteint des performances équivalentes ou supérieures aux contreparties classiques -- à savoir, une représentation en sacs de mots ou un modèle à champ aléatoire de Markov --, une première lors de la publication en 2013.

\subsection{Extensions du QLM}
\label{sec:ext_neuronales_qlm}

Plusieurs extensions du modèle QLM ont
été étudiées.
\citeasnoun{xie2015modeling} reprennent le QLM en modifiant la notion de dépendance. Plutôt que la simple cooccurrence, il s’agit des ensembles de termes statistiquement corrélés entre eux, au sens d’une équivalence à la propriété d’intrication quantique. Cela permet en pratique de mieux traiter les requêtes plus longues. Une explication avancée est que ce modèle élimine les cooccurrences redondantes avec les projecteurs individuels, qui introduisent un bruit inutile pour la requête.
\citeasnoun{li2018quantum} exploitent un algorithme de convergence globale du maximum de vraisemblance pour étendre la requête avec du vocabulaire hors requête.
\citeasnoun{zhang2018unsupervised} adaptent le modèle QLM sur la tâche de classification de sentiments.
Pour cela, ils créent artificiellement deux \frquote{dictionnaires de sentiments}, et représentent ces dictionnaires et les documents à classer par des matrices de densité. L'entropie quantique relative est ensuite utilisée pour mesurer la similarité entre les matrices de densité des documents et des dictionnaires. La faisabilité de l'approche est démontrée sur deux jeux de données provenant de Twitter en langue anglaise : \emph{Obama-McCain Debate (OMD)} et \emph{Sentiment Strength Twitter Dataset (SS-Tweet)}.

Le modèle \emph{Neural Network based Quantum-like Language Model} (NNQLM) de \citeasnoun{zhang2018end} étend quant à lui le QLM par deux aspects, sur une tâche de questions-réponses.
Premièrement, afin de prendre en compte des informations sémantiques globales, la matrice de densité correspondant à une phrase est construite à partir de vecteurs de plongement de mots $\mathbf{h}_w$ issus du modèle contextuel Word2Vec \textit{skip-gram} \cite{mikolov2013distributed}.
Un vecteur $\mathbf{h}_w$ est traité, après normalisation, comme l'observation d'un système quantique en état de superposition $\ket{w}=\mathbf{h}_w/||\mathbf{h}_w||_2$.
Une phrase $p=\{w_i\}_{i=1}^{n}$ est alors un mélange statistique de matrices de densité $\rho_p=\sum_i\theta_i\ket{w_i}\bra{w_i}, \sum_i\theta_i=1$ -- p. opp. à un encodage \emph{one-hot}, moins riche, et un état pur dans le QLM.
Deuxièment, un réseau convolutionnel en deux dimensions prend pour entrée la représentation jointe $\rho_q\rho_a$ d'une question $\rho_q$ et d'une réponse $\rho_a$, pour extraire automatiquement des caractéristiques de similarité entre chaque paire de documents à comparer.
Une couche neuronale dense avec activation \emph{softmax} est enfin ajoutée pour prédire si la réponse est correcte ou non.
Ainsi, NNQLM intègre QLM dans un réseau de neurones, et tire profit de données annotées grâce à une mise à jour par rétropropagation.
Par la suite, afin de répondre à une limite commune des modèles QLM et NNQLM, dont la représentation des phrases ne prend pas en compte l'ordre ou la position des mots (modèles en \frquote{sacs de mots}), \citeasnoun{zhang2022complex} ont étendu NNQLM avec un réseau de neurones quantique à valeurs complexes (C-NNQLM).
Le modèle est évalué sur des tâches (i) de questions-réponses (jeux de données TREC-QA, WIKIQA et YahooQA), (ii) de recherche de documents (données MS-MARCO), et (iii) de classification de textes (6 jeux de données contenant de 2 à 6 classes, des corpus de taille jusqu'à 120 k et 45 k mots de vocabulaire).
Les résultats sont supérieurs au modèle QLM, mais inférieurs aux réseaux de neurones classiques à l'état de l'art.

\subsection{Autres modèles de langue quantiques}

Un modèle de langue alternatif, qui n'utilise pas les matrices de densité, est étudié par \citeasnoun{zhang2018quantum}.
Dénommé QMWF-LM, \emph{Quantum Many-body Wave Function Language Model}, il repose sur une analogie avec le problème physique à $n$ corps et aboutit à un modèle de réseau de neurones convolutionnel, créant ainsi un lien théorique entre formalisme quantique et certaines architectures classiques.

Dans ce modèle, chaque mot d'un document est assimilé à une particule, qui vit dans son propre espace de Hilbert de dimension $m$.
En associant chacun des $m$ vecteurs de base à l'un des sens du mot, cette configuration modélise plus finement la polysémie:
un mot $w$ est représenté par $\ket{w}=\sum_{i=1}^{m}\alpha_{i}^{(w)}\ket{e_i^{(w)}}$.
La base $\{\ket{e_i^{(w}}\}_i$ de l'espace associé à $w$ est soit la base canonique (encodage \emph{one-hot}), soit construite à partir de plongements de mots normalisés, option retenue par les auteurs avec la représentation GloVe \cite{pennington2014glove}.
L'espace global -- \pex{} associé à une phrase $p=\{w_i\}_{i=1}^{n}$ -- est alors le produit tensoriel des $n$ espaces de mots. Il a pour dimension $m^n$ et est décrit par un vecteur d'état $\ket{\psi_S}=\ket{w_1}\otimes\dots\otimes\ket{w_n}$, appelé fonction d'onde.
Au-delà de cette représentation \emph{locale} d'une phrase, les auteurs cherchent une représentation \emph{globale} $\ket{\psi}$, qui dépend du corpus entier.
En projetant la représentation globale sur la représentation locale, \emph{via} le produit scalaire $\braket{\psi_S|\psi}$, ils obtiennent un score utilisable pour des tâches telles que la réponse à des questions.
Toutefois, la décomposition de la fonction d'onde globale $\ket{\psi}$ dépend d'un tenseur de très grande dimension, intractable.
La contribution de \citeasnoun{zhang2018quantum} est de montrer que ce tenseur se décompose en tenseurs de faible rang, et que cette décomposition correspond à une certaine architecture de réseaux de neurones convolutionnels, donnant un nouvel éclairage sur les connexions entre théorie quantique et apprentissage classique.
Des performances supérieures à QLM et à NNQLM sont obtenues sur le problème de questions-réponses sur les données TREC-QA, WIKIQA et YahooQA.
La comparaison à C-NNQLM est moins nette: QMWF-LM est au moins aussi bon sur les deux premiers jeux de données, mais nettement moins performant sur YahooQA, jeu de données en plus grande dimension (près de 60 k questions et 300 k paires questions-réponses contre environ 1 k questions pour les deux autres) \cite{zhang2022complex}.

{\`A} noter que la décomposition tensorielle est un outil utilisé \pex{} par \citeasnoun{ma2019tensorized} pour proposer une architecture moins gourmande en paramètres du mécanisme d'auto-attention des réseaux Transformers \cite{vaswani2017attention}, avec en prime de meilleures performances en modélisation de langue sur les jeux de données en grande dimension Penn Treebank et WikiTest-2\footnote{Code disponible : https://github.com/szhangtju/The-compression-of-Transformer.}. De même \citeasnoun{panahi2020wordket} ont proposé un modèle de compression de vecteurs de plongement de mots inspiré par l'intrication quantique, réduisant ainsi le nombre de paramètres entrainables.
Ce formalisme appartient à la théorie des \emph{réseaux de tenseurs}, dont les circuits quantiques sont un cas particulier \cite{biamonte2017tensor}.
\citeasnoun{zhang2019generalized} ont ainsi étendu le modèle QMWF-LM avec le modèle TSLM, \emph{Tensor Space Language Model}, qui apprend la probabilité conditionnelle $\mathbb{P}(w_t|w_{1:t-1})$ d'un mot $w_t$ sachant l'historique passé $w_1,\dots,w_{t-1}$ dans une séquence, à la façon des réseaux récurrents RNNs -- cf. section \ref{ssec:hybrid}.
Les auteurs montrent que TSLM généralise les équations des RNNs, et obtient de bons résultats par rapport à ces derniers sur Penn Treebank et \text{WikiTest-2}.
Cela illustre une nouvelle fois comment la théorie et les outils de l'informatique quantique
permettent de revisiter certaines architectures classiques de réseaux de neurones.

\subsection{Modélisation de relations lexicales}
\label{sec:modele_langue_ambig}

Une motivation de certains modèles de langue d'inspiration quantique est de représenter des relations lexicales, comme l'ambiguïté ou l'hyponymie.
En effet, d'après \citeasnoun{meyer2020modelling}, des modèles classiques contextuels représentant les mots par des vecteurs seuls peuvent gérer des cas de polysémie mais pas d'ambiguïté, telle l'homonymie.
Selon \citeasnoun{piedeleu2015open}, un mot polysémique a plusieurs sens reliés par un concept commun, tandis que deux homonymes ont des sens complètement distincts.
Ainsi, le mot \textit{bank} en anglais est polysémique, car il peut désigner une banque en tant qu'institution financière, ou le bâtiment où ces services sont délivrés.
Il est aussi homonyme, puisqu'il peut désigner les berges d'une rivière. 

\citeasnoun{piedeleu2015open} encodent un mot polysémique comme un état pur, et un homonyme comme un mélange statistique, avec une distribution de probabilité pour chaque sens.
Dans les deux cas, le mot est représenté par la matrice de densité associée $\rho$, et l'entropie de Von Neumann $\text{tr}(\rho\log\rho)$ est calculée pour mesurer l'ambiguïté, depuis un mot unique jusqu'à un constituant plus large, \pex{} de \emph{bank} à \emph{river bank}.
Chaque mot est initialisé avec un vecteur sémantique ayant pour base les 2000 mots pleins $W$ les plus fréquents. Les poids $v_i(t)$ du vecteur d'un mot cible $t$ dépendent des probabilités de chaque mot contextuel $c_i \in W$ d'apparaître dans le voisinage de $t$. Chaque vecteur est associé à un couple $(\text{lemme}, \text{catégorie grammaticale})$. Ainsi, le verbe et le nom anglais \emph{book} sont représentés par deux vecteurs différents.
Sur 5 mots ambigüs, les auteurs montrent que chaque mot seul a une mesure d'ambiguïté plus élevée que s'il est accompagné d'autres mots, ce qui valide la modélisation.

S'appuyant à la fois sur cette approche et sur le modèle contextuel Word2Vec \textit{skip-gram} avec échantillonage négatif (SGNS, \citeasnoun{mikolov2013distributed}), \citeasnoun{meyer2020modelling} développent le modèle de matrice de densité Word2DM. 
L'apprentissage par SGNS modifie les vecteurs de mots en maximisant la similarité des vecteurs des mots cooccurrents, et en minimisant celle de mots n'apparaissant pas dans le même contexte. 
L'algorithme est étendu pour produire des matrices de densité plutôt que des vecteurs, avec une fonction objectif spécifique pour préserver en particulier la propriété de positivité d'une matrice de densité \cite{meyer2020modelling}. Il repose sur l'apprentissage d'une matrice intermédiaire $B$ en calculant la matrice de densité $A = BB^\intercal$, exploitant la propriété que pour toute matrice $B$, le produit $BB^\intercal$ est semi-défini positif. La fonction objectif SGNS est modifiée de la façon suivante : $J(\theta)=\textup{log } \sigma(tr(A_tA_c))+\sum_{k=1}^{K}\textup{log }\sigma(-tr(A_tA_{w,k})))$, avec $A_t$ et $A_c$ les matrices de densité des mots cibles et de contexte, $A_{w,k}$ les matrices de densité de $K$ échantillons négatifs, et $\theta$ l'ensemble des poids des matrices intermédiaires $B_t, B_c$ et $B_{w,k}$.

Les résultats pour la version optimisée de Word2DM indiquent un bon encodage de l'ambiguïté, telle qu'évaluée d'après la corrélation entre le nombre de sens associés à ce mot (nombre de \emph{synsets} du mot sur l'ontologie WordNet \cite{miller1995wordnet}) et son entropie de Von Neumann, définie section \ref{sec:concepts}.

La représentation d'un mot par un opérateur positif -- en relâchant la contrainte de trace unitaire des matrices de densité -- est aussi employée par \citeasnoun{lewis2019modelling} pour prendre en compte l'hyponymie, en introduisant une relation hiérarchique entre les mots.
En effet, l'ensemble des opérateurs positifs d'un espace vectoriel est muni d'une relation d'ordre:
pour deux opérateurs $A$ et $B$, $A \sqsubseteq B \Leftrightarrow B - A$ est positif.
Soit un opérateur positif $\llbracket mammal\rrbracket$ représentant le mot \emph{mammifère} 
et un opérateur positif $\llbracket dog \rrbracket $ représentant le mot \emph{chien}, alors on obtient : $\llbracket dog \rrbracket \sqsubseteq \llbracket mammal \rrbracket $.
Un mot $w$ est alors vu comme une collection de vecteurs d'état $\ket{w_i}$ représentant chacun une instance du concept -- un hyponyme -- exprimé par $w$.
$\ket{w_i}$ est obtenu par un plongement de mots GloVe \cite{pennington2014glove}; les relations sont construites à l'aide de la base WordNet \cite{miller1995wordnet}, qui contient des relations d'hyponymie et d'hyperonymie entre mots de la langue anglaise.
Ainsi, pour chaque mot $w$, tous les hyponymes $w_i$ de $w$ à chaque niveau $i$ pour lequel il existe un vecteur GloVe sont collectés sur WordNet.
Finalement, $\llbracket w \rrbracket = \sum_i p_i \ket{w_i} \bra{w_i}$, où les poids $p_i$ sont dérivés du texte, sans normalisation. 
Les représentations obtenues sont testées sur trois jeux de données en anglais composés de phrases simples contenant ou non des implications \cite{sadrzadeh2018sentence}, ce qui est indiqué par un label positif ou négatif. Par exemple, {\og}l'été se termine, la saison prend fin{\fg} porte un label positif, mais {\og}la saison prend fin, l'été se termine{\fg} porte un label négatif. Les auteurs rapportent une bonne performance de leur modèle sur la tâche de détection d'implication.

\section{Modèles quantiques des langues}\label{sec:modeles}

Au-delà des approches décrites en section \ref{sec:matricesdensite}, des efforts ont été produits pour faire émerger des modèles spécifiquement quantiques de TAL.
Nous revenons en détail sur l'approche diagrammatique, visant à convertir les mots et leurs relations en états quantiques au moyen de symboles et de règles, et sur l'approche par réseaux de neurones hybrides, visant à transformer certaines parties des réseaux de neurones classiques en circuits quantiques.

\subsection{L'approche diagrammatique}

\subsubsection{Principe}

L'approche diagrammatique tire ses origines des travaux théoriques de \citeasnoun{coecke2010mathematical} d'inspiration quantique, définissant une représentation sémantique distributionnelle et compositionnelle du sens d'une phrase. Ils seront repris par \citeasnoun{zeng2016quantum} utilisant pour la première fois le calcul quantique appliqué au TAL, constituant les fondations du TQL. Le sens d'une phrase est calculé avec des principes mathématiques compatibles avec la physique quantique, à partir d'une représentation symbolique du lexique et de la syntaxe. Le modèle général est appelé DisCoCat d'après l'anglais \textit{DIStributional COmpositional CATegorical}.


DisCoCat emploie une grammaire catégorielle de prégroupe \cite{lambek1997type}. Dans cette grammaire, un type est associé à chaque mot, et des règles de réduction s'appliquent sur les types. Un type donné $p$ a un adjoint gauche $p^l$ et un adjoint droit $p^r$, et il existe deux règles de réduction : $p^l \cdot p \rightarrow 1$ et $p \cdot p^r \rightarrow 1$. Le type $n$ correspond aux noms et aux propositions nominales, le type $s$ aux phrases. Le type d'un verbe transitif est alors $n^r \cdot s \cdot n^l$, signifiant qu'un type $n$ est attendu à gauche ainsi qu'un autre à droite. La phrase est valide si les réductions successives, supprimant un type et son adjoint, aboutissent au type $s$.

Cette représentation grammaticale est ensuite convertie en diagramme de cordes, décrivant ici les relations grammaticales entre les mots sous forme de compositions séquentielles, admettant des entrées et produisant des sorties. De façon plus générale, les diagrammes de cordes correspondent à un langage graphique permettant de représenter et de manipuler les états quantiques. Les diagrammes expriment des calculs avec des catégories monoïdales, utilisées pour représenter des processus dans des systèmes de types variés, dont un ordinateur quantique. Une catégorie monoïdale est munie d'un bifoncteur généralisant la notion de produit tensoriel de deux structures algébriques. Cette notion est détaillée par \citeasnoun{coecke2010mathematical}. Un exemple de diagramme est donné sur la figure \ref{fig:lorenz_diag_and_circuit}.

Afin de réaliser des expérimentations, telles que celles présentées dans la section suivante \ref{sec:approches_diag_exp}, ces diagrammes sont convertis en circuits quantiques. Pour cela, il est nécessaire de fixer certains hyper paramètres définissant l'architecture du circuit : les nombres $q_n$ et $q_s$ de qubits associés à chaque corde respectivement de type $n$ et $s$, ainsi que les états quantiques avec lesquels tous les mots sont remplacés. Ces choix déterminent un \textit{ansatz} -- section \ref{sec:concepts} --, et sont fixés avant la phase d'entrainement. Les circuits sont entrainables : ils contiennent des paramètres libres, pouvant être optimisés pour la réalisation d'une tâche.

En complément, des travaux théoriques, considérés comme des extensions de DisCoCat par leurs auteurs, ont été proposés pour l'hyponymie \cite{lewis2019modelling} -- section~\ref{sec:modele_langue_ambig} --, et l'ambiguïté syntaxique \cite{correia2022quantum}. Pour illustrer ce type d'ambiguïté, les auteurs prennent pour exemple le groupe nominal {\og}\textit{rigorous mathematicians and physicists}{\fg} ({\og}des physiciens et des mathématiciens rigoureux{\fg}). L'adjectif {\og}rigorous{\fg} peut porter soit sur {\og}mathematicians and physicists{\fg}, soit sur {\og}mathematicians{\fg} uniquement. Afin de gérer ces deux représentations simultanément, elles sont placées sur un circuit commun, dans lequel des portes d'échange de deux qubits contrôlent la portée syntaxique. Un qubit $|c \rangle = c_1 | 1 \rangle + c_2 | 0 \rangle$ est ajouté afin de contrôler ces portes d'échange, avec les probabilités respectives $|c_1^2|$ et $|c_2^2|$ que le premier et le second sens se matérialisent.


\subsubsection{Travaux expérimentaux}
\label{sec:approches_diag_exp}

\citeasnoun{meichanetzidis2023grammar} réalisent la première expérimentation en TQL sur une machine NISQ. Ils instancient les phrases sur des VQCs au moyen du modèle DisCoCat. Le sens des mots est encodé par des états quantiques, et leurs relations de dépendance par des intrications. Les auteurs considèrent les types de la grammaire de prégroupe $n$ et $s$, et un nombre de qubits $q_n$ et $q_s$ est associé à chaque type. Ce nombre détermine l'arité $k$ de chaque mot $w$, \cad{} la largeur du circuit requise pour préparer l'état du mot. Les circuits correspondant au type $s$ sont des scalaires ($q_s =0$). Les mots unaires (1 qubit) sont représentés avec une paramétrisation d'Euler, \cad{} un circuit de décomposition des trois angles d'Euler $R_z(\theta_1) \circ R_x(\theta_2) \circ R_z(\theta_3)$. Ces angles peuvent décrire l'orientation d'un référentiel par rapport à un repère cartésien. Les mots avec une arité supérieure à 1 sont représentés par des circuits quantiques de type instantané à temps polynômial, en anglais \textit{Instantaneous Quantum Polynomial-time} (IQP) \cite{bremner2011classical}, comportant $d$ couches. Les portes d'un tel circuit sont commutatives car non dépendantes du temps, d'où le terme \frquote{instantané}. Un circuit IQP est constitué de portes d'Hadamard suivies d'une couche de portes de \text{Z-rotation} contrôlées $CR_z(\theta_i)$ telles que $i \in \{1,2,...,d(k -1)\}$. Le pronom relatif \textit{who} en anglais (pronom relatif sujet désignant une personne) est associé à un circuit GHZ, qui n'a pas de paramètres. Ce circuit est choisi d'après les travaux de \citeasnoun{sadrzadeh2013frobenius} sur les pronoms relatifs. Un état GHZ est une généralisation de l'état de Bell à trois qubits, représentant une intrication entre les qubits.

L'ensemble des $\theta$ est la paramétrisation du VQC associé à une phrase et son diagramme : ils sont optimisés de façon à résoudre le problème d'apprentissage formulé à partir de la mesure finale du circuit -- ici une classification binaire.

Ces circuits sont entrainés pour une tâche de questions-réponses, sur des données synthétiques créées pour l'expérimentation. L'emploi de données synthétiques permet d'obtenir des phrases courtes, en limitant les structures syntaxiques et le vocabulaire utilisé. Les paramètres du circuit sont entrainés pour prédire une réponse correcte (0 ou 1) à une question. Par exemple la question {\og}\textit{Romeo loves Juliet}{\fg} (Roméo aime Juliette) porte le label de réponse 1, et {\og}\textit{Romeo loves Romeo}{\fg} (Roméo aime Roméo) le label 0. Les auteurs réalisent une simulation classique, et une expérimentation sur machine NISQ. Pour la simulation classique, le corpus est de 30 phrases, avec un vocabulaire de 7 mots. Sur machine NISQ, le corpus est de 16 phrases et 6 mots de vocabulaire, soit le maximum des capacités du hardware utilisé. Les entrainements utilisent l'algorithme d'optimisation SPSA, un algorithme d'approximation du gradient \cite{spall1998implementation}.
Les auteurs testent 3 machines NISQ, et reportent les résultats de l'erreur d'entrainement et de test. L'erreur de test la plus basse est obtenue sur la machine \texttt{ibmq\_montreal} avec 0 en erreur d'entrainement, et 0,375 en erreur de test.

Avec la bibliothèque Python DisCoPy \cite{defelice2020discopy}, permettant de compiler des diagrammes en code, et donc d'utiliser le modèle DisCoCat, \citeasnoun{lorenz2021qnlp} proposent une expérimentation de classification de phrases sur des machines NISQ \footnote{Le code source des auteurs est à disposition sur https{:}//github.com/CQCL/qnlp\_lorenz\_etal\_2021\_resources.}.
Deux tâches sont abordées. La première est une tâche de classification binaire sur les thèmes de la nourriture et de l'informatique. Le jeu de données, appelé MC pour \textit{Meaning Classification}, contient 130 phrases d'au plus 5 mots. C'est un jeu synthétique généré automatiquement avec une grammaire indépendante du contexte (\textit{Context-Free Grammar, CFG} en anglais) comportant 4 règles de réécriture simples, et un vocabulaire fixe de 17 mots. Les données contiennent \pex{} les phrases {\og}\textit{skillful programmer creates software}{\fg} (le programmeur talentueux développe un logiciel) et
{\og}\textit{chef prepares delicious meal}{\fg} (le chef prépare un délicieux repas).

La seconde tâche est une classification visant à prédire si une proposition nominale contient une proposition relative référant à un sujet ou à un objet. Le jeu de données, appelé RP, contient 150 propositions nominales issues du corpus RELPRON \cite{rimell2016relpron}. La taille du vocabulaire est limitée à 115 mots, et chaque mot apparaît au moins 3 fois. Les données contiennent \pex{} les phrases : {\og}\textit{device that detects planets}{\fg} ({\og}instrument qui détecte les planètes{\fg}) et {\og}\textit{device that observatory has}{\fg} ({\og}instrument que l'observatoire possède{\fg}).

\begin{figure}[ht]
\begin{center}
    \includegraphics[width=1\columnwidth]{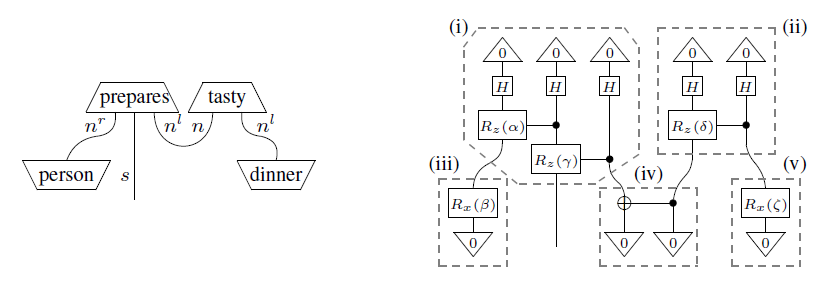}
\end{center}
\caption{Représentation de la phrase {\og}person prepares tasty dinner{\fg} (la personne prépare un diner savoureux) par un diagramme DisCoCat à gauche, et par un circuit quantique à droite \protect\cite{lorenz2021qnlp}. Les mots {\og}prepares{\fg} et {\og}tasty{\fg} sont remplacés respectivement par les états (i) et (ii), {\og}person{\fg} et {\og}dinner{\fg} respectivement par les effets (iii) et (v). La corde en forme de coupe entre {\og}prepares{\fg} et {\og}tasty{\fg} est traduite en intrication par effet de Bell dans le composant (iv). }\label{fig:lorenz_diag_and_circuit}
\end{figure}

Ces phrases sont transformées en diagrammes avec DisCoCat. Les représentations sont similaires à celles de l'expérimentation de \citeasnoun{meichanetzidis2023grammar} décrites plus haut.  
Un exemple fourni par les auteurs d'un diagramme DisCoCat et de sa transformation en VQC est montré en figure \ref{fig:lorenz_diag_and_circuit}.
 
Pour le paramétrage des circuits, le nombre de qubits fixé est le plus faible possible pour que l'expérimentation soit réalisable sur machine NISQ. Différentes configurations sont testées, en faisant varier les types d'effets représentant les noms, et la profondeur du circuit. Ces choix déterminent l'architecture du circuit, \cad{} l'\textit{ansatz}, et donc son nombre de paramètres.

Les auteurs rapportent en résultat un score F-Mesure de 0,85 et de 0,78 pour les tâches MC et RP respectivement, selon eux conforme à ce qui pourrait être attendu pour un jeu de données de cette taille. Ils ouvrent des perspectives sur l'utilisation de jeux de données plus larges (taille de vocabulaire, nombre de phrases), comptant sur une amélioration des capacités des machines NISQ.

\begin{figure}[ht]
	\begin{center}
		\includegraphics[width=1\columnwidth]{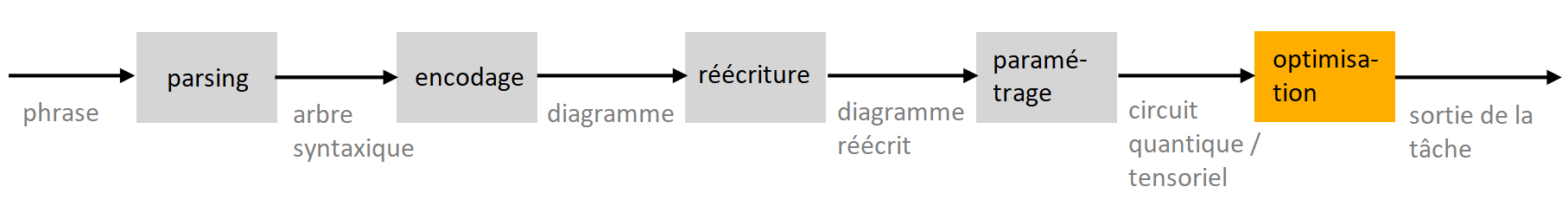}
	\end{center}
	\caption{Traitement d'une phrase avec \texttt{lambeq} \protect\cite{kartsaklis2021lambeq} } \label{fig:pipeline_lambek}
\end{figure}

Afin de faciliter les expérimentations en TQL reposant sur l'approche diagrammatique, la bibliothèque Python \texttt{lambeq} \cite{kartsaklis2021lambeq} a été développée. Elle est maintenant intégrée à la bibliothèque PennyLane pour l'informatique quantique.
Le lecteur intéressé pourra trouver des tutoriels sur les sites des bibliothèques concernées. \texttt{lambeq} permet d'encoder une phrase vers un circuit quantique selon le modèle DisCoCat, et d'optimiser les paramètres du circuit pour une tâche donnée. La bibliothèque DisCoPy \cite{defelice2020discopy} est utilisée pour le chargement des diagrammes et pour leur manipulation. Pour l'encodage, \texttt{lambeq} fournit notamment pour l'anglais un parseur à l'état de l'art obtenu par apprentissage automatique \cite{yoshikawa2017ccg}. Il est fondé sur une grammaire catégorielle combinatoire (GCC, ou CCG en anglais pour \textit{Combinatory Categorial Grammar}) \cite{steedman2001syntactic}. Celle-ci a été préférée à la grammaire de prégroupe pour son fort pouvoir d'expressivité, et parce qu'il existait déjà des parseurs CCG automatisés. Un schéma de la chaîne de traitement d'une phrase par la bibliothèque \texttt{lambeq} est donné en figure \ref{fig:pipeline_lambek}.

\subsection{Hybridation des réseaux de neurones} \label{ssec:hybrid}

Nous avons abordé section \ref{sec:ext_neuronales_qlm} les approches d'inspiration quantique, avec un premier travail en 2018 \cite{zhang2018end} sur une extension du modèle QLM. Il existe un second type d'approche fondé sur des réseaux de neurones hybrides, conçus pour être exécutés sur un ordinateur quantique ou sur une machine NISQ. Il utilise à la fois des VQCs et des étapes de calcul classique au sein d'une même architecture. Son objectif est de tirer parti  de la puissance des réseaux de neurones profonds classiques déjà démontrés pour le TAL, et des avantages théoriques de l'informatique quantique.

Les architectures à longue mémoire à court terme (en anglais \textit{Long Short-Term Memory}, LSTM) \cite{hochreiter1997long} et Transformer \cite{vaswani2017attention} ont ainsi été adaptées en réseaux de neurones hybrides.

Un réseau de neurones LSTM est une extension d'un réseau de neurones récurrent (RNN), mentionné section \ref{sec:ext_neuronales_qlm}. Un RNN possède un état caché $h_t$ étant à la fois une entrée et une sortie du réseau qui l'actualise à chaque temps $t$. Cet état représente la mémoire permettant de transmettre les informations des étapes précédentes. Une limite des RNN est qu'ils ne peuvent pas transmettre des informations sur une longue distance en raison du pénomène de disparition du gradient. L'architecture LSTM pallie cette limite en intégrant une cellule de mémoire supplémentaire $c_t$, actualisée au temps~$t$, permettant au gradient de persister dans le réseau.

\citeasnoun{chen2022quantum} conçoivent un modèle LSTM hybride quantique classique appelé QLSTM. Le modèle est capable d'apprendre sur plusieurs types de données temporelles sur des problèmes physiques. Dans certains cas, les performances du QLSTM sont supérieures à sa contrepartie classique, ou convergent plus rapidement. \citeasnoun{abbaszade2021application} se fondent sur cet avantage pour proposer un algorithme pouvant traduire des phrases de l'anglais vers le persan avec un QLSTM. Les exemples traités sont des phrases simples avec une structure grammaticale comprenant un sujet, un verbe transitif, et un objet. Les phrases sont encodées en circuit avec DisCoCat selon la méthode décrite par \citeasnoun{lorenz2021qnlp}. Les caractéristiques du circuit de la phrase à traduire sont passées en entrée à un QLSTM, qui les encode vers un vecteur caché appris, et les décode vers un vecteur de sortie correspondant à la phrase suivante, \cad{} la phrase traduite. Un QLSTM contient six VQCs, qui remplacent certaines opérations du LSTM classique.
Chaque VQC est composé de trois couches : une non paramétrique d'encodage des données sous forme de qubits, une pour l'optimisation des paramètres (apprentissage des \textit{ansätze}), une pour la mesure. Nous détaillons ci-dessous pour comparaison les équations du LSTM classique (colonne gauche) et du QLSTM (colonne droite), avec $v_t = [h_{t-1}, x^t]$, la concaténation des états cachés $h$ au temps $t-1$ et des entrées $x$ au temps $t$, $*$ l'opérateur de multiplication terme à terme et $\sigma$ la fonction sigmoïde.
\begin{align}
f_t &= \sigma(W_f \cdot v_t + b_f) & f_t &= \sigma(VQC_1(v_t))
\label{eq:lstmvqc_forget}\\
i_t &= \sigma(W_i \cdot v_t + b_i) & i_t &= \sigma(VQC_2(v_t))
\label{eq:lstmvqc_relevant_inputs}\\
\tilde{C}_t &= tanh(W_c \cdot v_t + b_C) & \tilde{C}_t &= tanh(VQC_3(v_t))
\label{eq:lstmvqc_cell_pre_update}\\
c_t &= f_t * c_{t-1} + i_t * \tilde{C}_t & c_t &= f_t * c_{t-1} + i_t * \tilde{C}_t
\label{eq:lstmvqc_cell_update}\\
o_t &= \sigma(W_o \cdot v_t + b_o) & o_t &= \sigma(VQC_4(v_t))
\label{eq:lstmvqc_relevant_values}\\
h_t &= o_t * tanh(c_t) & h_t &= VQC_5(o_t * tanh(c_t))
\label{eq:lstmvqc_hidden_output}\\
 & &y_t &= VQC_6(o_t * tanh(c_t))
\label{eq:lstmvqc_y_output}
\end{align}

La porte d'oubli $f_t \in [0,1]$ (Eq. \eqref{eq:lstmvqc_forget}) détermine dans quelle mesure les éléments correspondants dans l'état de la cellule $c_{t-1}$ doivent être oubliés ou mémorisés (par application de $f_t * c_{t-1}$ Eq. \eqref{eq:lstmvqc_cell_update}). La porte d'entrée $i_t$ détermine quelles valeurs seront ajoutées à l'état de la cellule (Eq.\eqref{eq:lstmvqc_relevant_inputs}). $\tilde{C}_t$ est l'état de la cellule candidate (Eq. \eqref{eq:lstmvqc_cell_pre_update}) utilisé pour la mise à jour de l'état de la cellule $c_t$ (Eq. \eqref{eq:lstmvqc_cell_update}). Après la mise à jour de $c_t$, les sorties peuvent être calculées. $o_t$ représente les valeurs de l'état de la cellule $c_t$ pertinentes pour la sortie (Eq. \eqref{eq:lstmvqc_relevant_values}). Pour le QLSTM, $o_t$ et $c_t$ sont utilisés pour calculer l'état caché $h_t$ avec le $VQC_5$, et la sortie $y_t$ avec le $VQC_6$ (Eq. \eqref{eq:lstmvqc_hidden_output} et \eqref{eq:lstmvqc_y_output}).


\citeasnoun{di2022dawn} proposent une architecture alternative de LSTM hybride, dans laquelle le VQC est positionné entre deux couches classiques. Un VQC ne pouvant pas changer les dimensions des données d'entrée, la première couche adapte la dimensionalité au nombre de qubits du VQC, et la seconde adapte la dimensionalité de la sortie du VQC pour les vecteurs cachés.
Le modèle est implémenté sur une tâche d'étiquetage morpho-syntaxique (\textit{Part-Of-Speech tagging})\footnote{Le code source des auteurs est à disposition sur https{:}//github.com/rdisipio/qlstm.}, exécutée en simulation quantique, \cad{} sur machine classique. Le jeu de données constitué à la main est composé de deux phrases simples en anglais : {\og}\textit{The dog ate the apple}{\fg} (le chien a mangé la pomme) et {\og}\textit{Everybody read that book}{\fg} (tout le monde a lu ce livre). Le jeu de test est identique au jeu d'entrainement. Les résultats des modèles hybrides et classiques sont similaires, avec 100~\% de résultats corrects. Le LSTM hybride doit être entrainé plus longtemps en simulation que sa contrepartie classique (15 minutes contre 8 secondes), mais utilise deux fois moins de paramètres (199 contre 477).

Les auteurs proposent également une architecture de Transformer hybride. Un modèle Transformer classique \cite{vaswani2017attention} gère les dépendances longues dans les textes grâce à un mécanisme d'auto-attention prenant en entrée des représentations de plongements de mots. 
Une couche d'auto-attention possède trois types de vecteurs : un vecteur requête $Q$, un vecteur clé $K$, et un vecteur valeur $V$, chacun associé à une matrice de poids entrainable. Chaque vecteur est calculé en multipliant les vecteurs de plongement de mots en entrée par la matrice entrainée correspondante.
Soit $d_k$ la taille des requêtes et des clés, les sorties de la couche d'attention sont calculées par la formule : $Attention(Q, K, V) = softmax(\frac{QK^T}{\sqrt{d_k}})V$.
Pour concevoir un modèle Transformer hybride, \citeasnoun{di2022dawn} remplacent les transformations linéaires du mécanisme d'attention par des VQCs. Ils réalisent une expérimentation quantique simulée avec cette architecture sur une tâche de classification de sentiments sur le corpus IMDB\footnote{Code source à disposition sur https://github.com/rdisipio/qtransformer.}. Les scores de performance ne sont pas rapportés. L'entrainement d'une seule itération sur tout le jeu de données -- environ cent heures -- s'est révélé trop long pour optimiser pleinement les paramètres du modèle. 

Une approche similaire fondée sur un Transformer a été développée par \citeasnoun{li2022quantum}, avec une architecture de réseau de neurones à attention quantique \textit{quantum self-attention neural network} (QSANN) utilisant une projection gaussienne de l'auto-attention. D'après les auteurs, les travaux de \citeasnoun{di2022dawn} utilisent directement le produit scalaire dans l'auto-attention, ce qui ne permet pas d'établir des corrélations entre des éléments lointains. La projection gaussienne utilisée dans un QSANN exploite un espace de Hilbert exponentiellement large, à même de prendre en compte des corrélations cachées entre les mots, ce qui pourrait permettre de dépasser les performances des architectures classiques.



\begin{figure}[ht]
	\begin{center}
		\includegraphics[width=1\columnwidth]{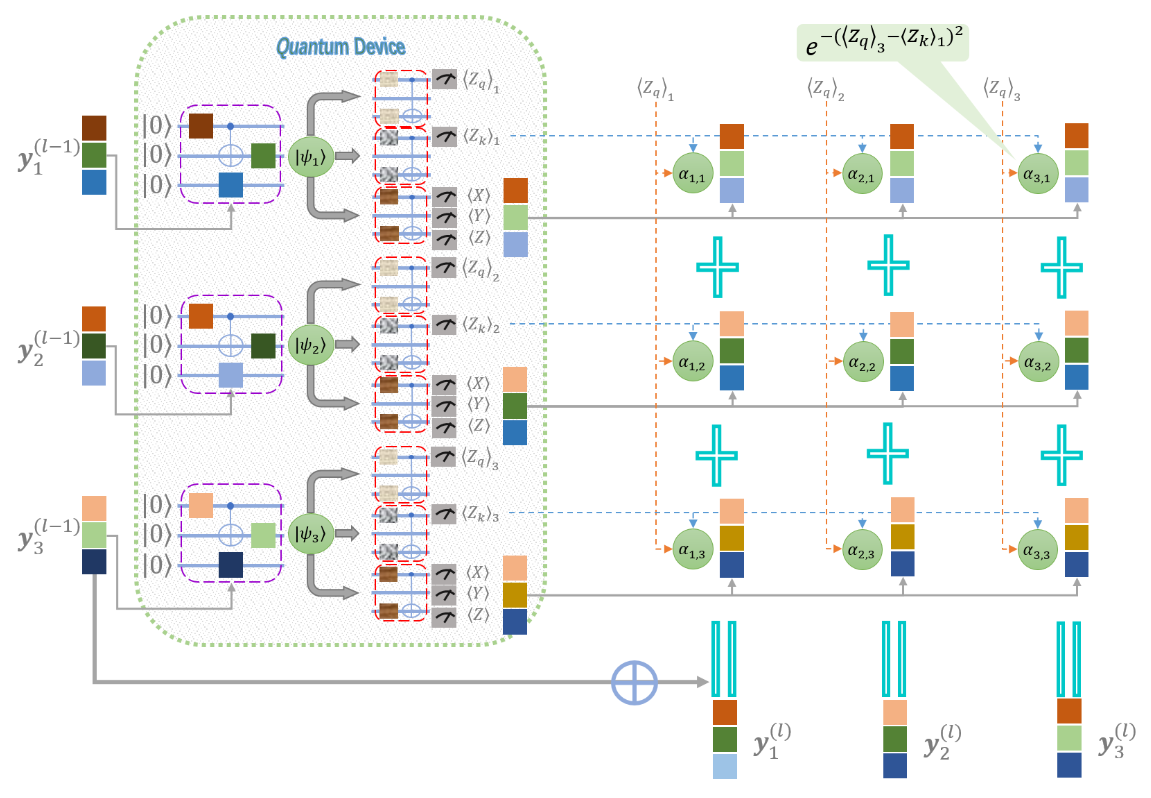}
	\end{center}
	\caption{Schéma d'une couche d'auto-attention quantique par \protect\citeasnoun{li2022quantum} } \label{fig:QSAL}
\end{figure}

Dans ce modèle, la couche d'auto-attention d'un Transformer classique est remplacée par une couche d'auto-attention quantique, dénommée \emph{Quantum Self Attention Layer} (QSAL). Le schéma de cette couche est présenté sur la figure \ref{fig:QSAL}. Sur machine quantique, les entrées $\{y_s^{(l-1)}\}$ de la couche $l$ sont utilisées comme angles de rotation d'\textit{ansatz} (boîtes en ligne pointillée rouge) pour les encoder dans leur état quantique $\{|\psi_s \rangle\}$. Ensuite pour chaque état, trois classes d'\textit{ansatz} sont exécutées : les deux premières correspondent à la requête et à la clé, la troisième correspond à la valeur. Les sorties des requêtes $\langle Z_q \rangle_s$ et des clés $\langle Z_k \rangle_j$ sont mesurées et calculées sur machine classique par une fonction gaussienne pour obtenir les coefficients d'auto-attention $\alpha_{s,j}$ (cercles verts). Toujours sur machine classique, la somme pondérée par $\alpha_{s,j}$ de la partie correspondant aux valeurs (troisième \textit{ansatz}, petits carrés colorés) est calculée, puis les entrées $y$ sont ajoutées pour obtenir les sorties $\{y_s^{(l)}\}$.

Les auteurs évaluent leur modèle sur deux tâches. La première est une classification thématique identique à celle réalisée par \citeasnoun{lorenz2021qnlp} avec l'outil DisCoCat au moyen d'une approche diagrammatique. Les résultats obtenus sont équivalents entre le QSANN et le modèle diagrammatique. La seconde tâche est une classification binaire de sentiments sur trois jeux de données issus du monde réel, contenant des commentaires sur des restaurants, des films et des produits, sélectionnés respectivement dans Yelp, IMDb et Amazon. Chaque jeu contient 1000 séquences avec autant d'avis négatifs que positifs, dont la longueur varie de quelques mots à plusieurs dizaines de mots.
QSANN se révèle un peu plus performant que sa contrepartie classique sur les trois jeux de données, avec moins de paramètres (785 contre 49).

\citeasnoun{yang2022bert} proposent le premier modèle hybride avec un Transformer BERT \cite{devlin2019bert} pour une tâche de classification d'intention. BERT est un modèle de langue général pouvant être adapté à une tâche spécifique, entrainé en masquant partiellement des entrées que le modèle doit apprendre à prédire. Les auteurs proposent une architecture composée d'un modèle BERT préentrainé, suivi d'un décodeur quantique, qui prend en entrée les plongements de mots issus du modèle BERT. Les données classiques sont transférées sur circuit quantique au moyen d'une convolution temporelle quantique, utilisée dans des travaux sur le traitement de l'image \cite{henderson2020quanvolutional} et de la voix \cite{yang2021decentralizing}. Cette technique découpe le signal en fenêtres glissantes traitées par un filtre convolutionnel temporel. Les représentations latentes ainsi obtenues après filtrage sont fournies au VQC. La taille des vecteurs de plongement de mots du modèle BERT est limitée à 50, et la taille de la fenêtre glissante est de 4, ce qui détermine le nombre de qubits du circuit. 

Les expérimentations sont réalisées en simulation quantique et sur machine NISQ avec deux jeux de données sur une tâche de classification d'intention : Snips \cite{coucke2018snips}, contenant des phrases prononcées dans un contexte d'interaction vocale avec un assistant personnel, et ATIS (AirLine Travel Information Systems) \cite{hemphill1990atis}, contenant des phrases prononcées correspondant à des réservations de vol en avion. Le modèle est testé sur les données texte avec différentes configurations sur le nombre et la taille des filtres. Les meilleurs scores pour les jeux Snips et ATIS sont respectivement de 96,62~\% et de 96,98~\% d'exemples correctement classés. Ces scores sont inférieurs à ceux pouvant être obtenus actuellement avec un modèle classique fondé sur une architecture Transformer, comme le montrent les résultats reportés par \citeasnoun{rafiepour2023ctran}, avec respectivement 99,42~\% et 98,07~\% d'intentions correctement détectées pour Snips et ATIS. Ces travaux classiques incorporent le modèle BERT large, dont la taille des vecteurs de plongement de mots est de 1024. En comparaison, pour l'expérimentation quantique de \citeasnoun{yang2022bert} la taille du modèle BERT est limitée à 50.

\section{Synthèse des travaux et discussion} \label{sec:discussion}

Nous avons présenté dans cet état de l'art des modèles d'inspiration quantique destinés à être exécutés sur machine classique, ainsi que des modèles quantiques conçus pour des simulations quantiques ou des machines NISQ.
Le tableau \ref{tab:synthese} liste ces modèles quantiques, ainsi que les tâches et les données sur lesquelles ils ont été testés. Les expérimentations réalisées uniquement avec les approches diagrammatiques sont actuellement limitées à des phrases synthétiques, tandis qu'il est déjà possible d'expérimenter sur des données naturelles avec des réseaux de neurones hybrides. 


\begin{table}[ht]
	\small
	\begin{center}
		\begin{tabular}{|r|c|c|c|}
			\hline
			\rule{0pt}{12pt}{Modèle} & Tâches & Données & Moyen \\[2pt]
			\hline
			\hline
			\rule{0pt}{12pt}DisCoCat & Questions-réponses & Synth. & Simulation \\
			\cite{meichanetzidis2023grammar}  &  &  & NISQ \\[2pt]
			\hline
			DisCoCat & Classification thématique & Synth. (\scriptsize{MC}) & NISQ \\
            \cline{2-3} 
			\cite{lorenz2021qnlp}  & Résolution d'anaphore & Synth. (\scriptsize{RP}) &  \\[2pt]
			\hline
			LSTM hybride & Etiquetage  & Synth. & Simulation   \\
			\cite{di2022dawn}  &  morpho-syntaxique &  &  \\[2pt]
			\hline 
			Transformer hybride & Classification de sentiment & Naturelles & Simulation \\
			\cite{di2022dawn} &   & \scriptsize{IMDb} &  \\[2pt]
			\hline
			Transformer hybride & Classification thématique & Synth. (\scriptsize{MC}) & Simulation \\
			\cline{2-3}
			QSANN & Résolution d'anaphores & Synth. (\scriptsize{RP}) &  \\[2pt]
			\cline{2-3}
			\cite{li2022quantum} & Classification de sentiment  & Naturelles &  \\
			&   & \scriptsize{Yelp IMDb Amazon} &\\[2pt]
			\hline
			Transformer hybride & Classification d'intention & Naturelles. & Simulation  \\
			BERT-QTC  &  & \scriptsize{Snips, ATIS} & NISQ  \\[2pt]
            \cite{yang2022bert} &  & & \\[2pt]
			\hline
		\end{tabular}
	\end{center}
	\caption{Principaux modèles en TQL exécutables sur machine NISQ et / ou en simulation quantique. La colonne {\og}Données{\fg} indique si les données utilisées sont synthétiques, ou naturelles.}\label{tab:synthese}
\end{table}

Le recours aux données synthétiques pour les approches diagrammatiques est nécessaire afin de construire des phrases courtes en limitant leur vocabulaire et leur syntaxe. \citeasnoun{lorenz2021qnlp} reportent qu'au moment de leur expérimentation, les capacités des machines NISQ n'ont pas permis d'augmenter davantage la longueur des phrases et la taille du jeu de données. Ces approches nécessitent également des étapes de prétraitement pouvant s'avérer coûteuses en temps, comme l'étiquetage morpho-syntaxique des phrases, et l'ajustement des circuits quantiques, afin que ces derniers puissent représenter le vocabulaire et la syntaxe d'une phrase tout en étant exécutables sur une machine NISQ. 
En pratique, des formes d'\textit{ansatz} sont choisies de par leur disponibilité sur la machine quantique utilisée \cite{lorenz2021qnlp}.
{\`A} l'inverse, les approches par réseaux de neurones hybrides utilisent les représentations par plongements de mots, permettant de prendre en compte des phrases complexes issues de jeux de données naturels. Pour autant, ces réseaux de neurones hybrides se confrontent actuellement à des temps d'exécution élevés.

Certains chercheurs \cite{li2022quantum} défendent l'approche par réseaux de neurones hybrides car elle permet de prendre en compte toute la complexité de la langue sans recourir à un prétraitement des données. D'autres \cite{correia2022quantum} souhaitent au contraire développer une approche fondée sur les règles de grammaire en TAL, pour privilégier des développements purement quantiques. Les travaux présentés dans cet état de l'art montrent que ces deux approches peuvent être complémentaires. En effet, l'approche diagrammatique telle qu'implémentée par la bibliothèque \texttt{lambeq} \cite{kartsaklis2021lambeq} utilise un étiquetage grammatical CCG dont la catégorie lexicale est prédite grâce à un réseau de neurones BERT \cite{devlin2019bert}, et le modèle LSTM hybride proposé par \citeasnoun{abbaszade2021application} encode les phrases avec DisCoCat, un formalisme appartenant à l'approche diagrammatique.

La diversité des modèles présentés montre que le dévelopement actuel des machines quantiques permet de réaliser des premières expérimentations, et de travailler sur de nouvelles architectures et de nouveaux algorithmes sur la base de ces résultats. Ces expérimentations restent limitées par les capacités actuelles des machines quantiques. Les gains espérés par les modèles quantiques en TQL, comme une meilleure gestion de l'ambiguïté \cite{meyer2020modelling}, la meilleure prise en compte des dépendances longues \cite{li2022quantum}, ou une accélération du temps de traitement, sont théoriquement fondés, mais ces avantages restent à confirmer.

Concernant les gains espérés sur la complexité algorithmique, \citeasnoun{wiebe2015quantum} présentent une variante quantique de l'algorithme du plus proche voisin, démontrant théoriquement un gain quadratique pour ce problème sous certaines conditions. \citeasnoun{zeng2016quantum} proposent une adaptation de cet algorithme dans le cadre du modèle DisCoCat, qui conserverait cet avantage. \citeasnoun{wiebe2019quantum} développent une représentation des structures syntaxiques pouvant conduire à un avantage exponentiel sur l'estimation de la validité grammaticale d'une phrase.

Toutefois, en raison du faible nombre de qubits disponibles, les performances en simulation quantique ou sur machine NISQ demeurent inférieures à celles des modèles classiques. Par exemple, pour l’expérimentation quantique de \citeasnoun{yang2022bert}, la taille des vecteurs de plongement de mots du modèle BERT est limitée à 50, alors qu'elle est de 1024 dans les expérimentations classiques, dont les scores surpassent ceux du modèle quantique.
Au sujet de leur expérimentation sur machine NISQ, \citeasnoun{lorenz2021qnlp} ont pour objectif explicite de rendre compte de leur démarche et des résultats obtenus afin de les partager à la communauté TAL, mais sans chercher à prouver un avantage quantique sur le temps de calcul en raison des capacités limitées des machines quantiques. Des ordinateurs quantiques de plus grande capacité seraient nécessaires pour évaluer pleinement le potentiel des modèles quantiques.

En outre, à l'ère du NISQ, une autre limitation pratique apparaît lors de l'optimisation des VQCs, même pour des problèmes de petite taille.
Typiquement, ces circuits sont optimisés \emph{via} un algorithme de descente de gradient, \pex{} SPSA \cite{spall1998implementation}, ce qui reste un défi pour deux raisons. 

Tout d'abord, des chercheurs ont mis en évidence le phénomène de \emph{plateau aride} (en anglais: \emph{barren plateau}), consistant en l'effacement exponentiel du gradient dans toutes les directions respectivement à la taille du système quantique \cite{mcclean2018barren}.
De même, \citeasnoun{holmes2022connecting} montrent qu'augmenter la capacité des \textit{ansätze} peut réduire le gradient.
Ainsi, le plateau aride aboutit au paradoxe que l'augmentation du nombre de qubits des machines NISQ peut rendre le problème d'optimisation plus compliqué et fournir de moins bonnes solutions.
Ce phénomène est observé en TQL par \citeasnoun{niroula2022constrained} sur un problème de résumé extractif de texte, avec de moins bonnes performances de l'algorithme quantique L-VQE lorsque le nombre de qubits augmente de 14 à 20.
La mise au point de stratégies pour contrer les plateaux arides est un pan important de la recherche sur les VQCs, \pex{} avec de meilleures initialisations des paramètres \cite{grant2019initialization}.

Le second défi pratique du NISQ porte sur le calcul même du gradient.
Les auteurs ont recours soit comme \citeasnoun{chen2022quantum}, à une expression analytique du gradient disponible pour certaines classes d'\textit{ansatz} \cite{schuld2019evaluating,crooks2019gradients}, soit comme \citeasnoun{meichanetzidis2023grammar}, à un schéma  numérique d'approximation du gradient.
Les deux cas requièrent d'évaluer au moins deux fois la fonction objectif, \text{\cad{}} le circuit quantique. La mise au point d'algorithmes d'optimisation robustes, au meilleur compromis entre la qualité et le coût de l'estimation, et adaptés à ces environnements stochastiques est donc une condition supplémentaire pour évaluer pleinement les concepts proposés en TQL.

\section{Conclusion et perspectives} \label{sec:conclusion}

Les travaux présentés dans cet état de l'art combinent le TAL et l'informatique quantique avec deux objectifs principaux : réaliser des tâches en obtenant de meilleures performances par rapport aux modèles dits {\og}classiques{\fg}, et mieux modéliser des phénomènes linguistiques comme l'ambiguïté et les dépendances longues.
Les méthodes utilisées spécifiques à l'informatique quantique, telles que les VQCs, ont motivé la conception de nouvelles approches. L'approche symbolique, dite {\og}diagrammatique{\fg}, permet de convertir un texte en circuit quantique, et l'approche par réseaux de neurones hybrides remplace certaines parties de réseaux classiques par des circuits quantiques. Les phénomènes quantiques de superposition et d'intrication sont aussi exploités par certains modèles. La spécificité de ces méthodes est à l'origine d'un nouveau champ disciplinaire : le traitement quantique des langues.

Les machines quantiques ont actuellement une capacité de calcul limitée en raison du faible nombre de qubits disponibles. Malgré cela, des expérimentations sur machines NISQ ont pu être réalisées pour quelques modèles. Même si les gains de performance sont encore difficiles à démontrer dans ces conditions contraintes, cela montre qu'il est déjà possible de concevoir des modèles quantiques en TAL, et de travailler sur l'architecture et les opérations de modèles quantiques. Les futurs travaux de recherche pourraient porter sur des extensions de modèles existants, et sur l'étude de nouvelles tâches. Par exemple \citeasnoun{di2022dawn} mentionnent que tous les blocs d'un réseau de neurones Transformer pourraient être remplacés par des VQCs, et que le modèle hybride pourrait être utilisé pour la génération automatique des langues. Il existe également des perspectives concernant le développement du TQL pour la langue française. Nous n'avons pas recensé à ce jour de travaux en TQL réalisés sur des données en français, et certaines ressources ne sont pas disponibles pour le français. Par exemple, le parseur BobCat utilisé dans la bibliothèque \texttt{lambeq} \cite{kartsaklis2021lambeq} conçue pour convertir des phrases en circuits quantiques gère uniquement l'anglais et l'allemand. Enfin, il n'existe pas à notre connaissance de jeu de données commun, ou de procédure d'évaluation commune, pour comparer les modèles quantiques. Construire un tel jeu de données serait bénéfique pour les futurs développements dans ce domaine.



\bibliography{qnlp}

\end{document}